\documentclass[conference]{IEEEtran}
\IEEEoverridecommandlockouts
\usepackage{cite}
\usepackage{amsmath,amssymb,amsfonts}
\usepackage{algorithmic}
\usepackage{graphicx}
\usepackage{textcomp}
\usepackage{xcolor}
\def\BibTeX{{\rm B\kern-.05em{\sc i\kern-.025em b}\kern-.08em
    T\kern-.1667em\lower.7ex\hbox{E}\kern-.125emX}}
\begin{document}

\title{Real-Time Anomaly Detection in Data Centers \\ for Log-based Predictive Maintenance using an Evolving Fuzzy-Rule-Based Approach}

\author{
\IEEEauthorblockN{Leticia Decker}
\IEEEauthorblockA{\textit{University of Bologna}\\
INFN Bologna \\
Bologna, Italy \\
leticia.deckerde@unibo.it}
\and
\IEEEauthorblockN{Daniel Leite}
\IEEEauthorblockA{\textit{Federal University of Lavras}\\ Department of Automatics\\
Lavras, Brazil \\
daniel.leite@ufla.br}
\and
\IEEEauthorblockN{Luca Giommi}
\IEEEauthorblockA{\textit{University of Bologna}\\
INFN Bologna \\
Bologna, Italy \\
luca.giommi3@unibo.it}
\and
\IEEEauthorblockN{Daniele Bonacorsi}
\IEEEauthorblockA{\textit{University of Bologna}\\
INFN Bologna \\
Bologna, Italy \\
daniele.bonacorsi@unibo.it}
}

\maketitle

\begin{abstract}

Detection of anomalous behaviors in data centers is crucial to predictive maintenance and data safety. With data centers, we mean any computer network that allows users to transmit and exchange data and information. In particular, we focus on the Tier-1 data center of the Italian Institute for Nuclear Physics (INFN), which supports the high-energy physics experiments at the Large Hadron Collider (LHC) in Geneva. The center provides resources and services needed for data processing, storage, analysis, and distribution. Log records in the data center is a stochastic and non-stationary phenomenon in nature. We propose a real-time approach to monitor and classify log records based on sliding time windows, and a time-varying evolving fuzzy-rule-based classification model. The most frequent log pattern according to a control chart is taken as the normal system status. We extract attributes from time windows to gradually develop and update an evolving Gaussian Fuzzy Classifier (eGFC) on the fly. The real-time anomaly monitoring system has to provide encouraging results in terms of accuracy, compactness, and real-time operation.

\end{abstract}


\section{Introduction}

A computing center (CC) is responsible for supporting a flexible, on-demand, dynamic, and computing-scalable cloud infrastructure, in which the resources are available directly or by means of services \cite{Escalante}. The complex CC infrastructure requires maintenance tools to keep itself operative, efficient, and reliable. 

The maintenance of a CC is based on the complexity of the operation and idling time. It is usually classified as: (i) reactive; (ii) preventive; (iii) predictive; and (iv) advanced. The reactive maintenance refers to a set of procedures deployed after the fault occurrence, which aims at restoring the pristine behavior. Preventive maintenance is the collection of procedures performed to lessen the likelihood of a system failure. The predictive maintenance is designed to determine the status of running services, and predict events of interest. Advanced maintenance combines the other three paradigms in order to forecast and diagnose failures \cite{16_Trojan}. 

Usually, CC maintenance is based on offline statistical analysis of log records -- in the preventive case, this is based on fixed time intervals. Recently, online computational-intelligence-based systems, namely, evolving fuzzy modelling frameworks \cite{Skrjanc1} \cite{Cordovil} \cite{Casalino} \cite{Garcia} \cite{Hyde} \cite{petronio2} supported by fast incremental machine-learning algorithms, have been employed in general issues related to on-demand anomaly detection, forecasting, autonomous data classification, and predictive maintenance of a plethora of applications \cite{Venkatesan} \cite{lobo} \cite{Angelov} \cite{Pratama1} \cite{EAISlog} \cite{Decker2}.

Log records concern service-oriented unstructured data. Log data samples need to be ad-hoc processed by learning and modelling algorithms. The use of general-purpose solutions based on the content of log files has been a challenge over the years. In a log-based system, the data may be highly verbose such that it is hard to extract useful information from raw data. The amount of data is huge while a high percentage tends to be redundant. Any CC service run by a user generates log data using multiple files. After being processed, a reasonable amount of data for analysis is obtained.

Since all CC activities are recorded in log files, algorithms can track event occurrences through the data extracted from log files to monitor and predict the system status. In the predictive case, identification of anomalous behavior as an intermediate step using global attributes of log records is possible \cite{Decker}. To reduce log-content processing, a common characteristic of the log files -- the timestamp of each line writing -- can be used. Furthermore, a reasonable assumption is that the system activity is proportional to the per-minute rate of lines written in a log file. Considering overall system faults, a direct impact on such rate of written log records is expected. 

The background scenario of this study is the Tier-1 Bologna -- the main Italian WLCG (Worldwide LHC Computing Grid) tier hold by the computing center of the Italian Institute of Nuclear Physics (INFN-CNAF). The WLCG involves 170 computing centers in over 42 countries, being the grid system that support the physics experiments performed at the biggest particle accelerator in the world, the CERN. It is organised in 4 layers -- the tiers, from 0 (at CERN) to 3, in decreasing order of importance. The Tier-1 Bologna has approximately 40,000 CPU cores, 40 PB of disk storage, and 90 PB of tape storage. It is connected to the Italian (GARR) and European (GÉANT) research networks, whose bandwidth of data transmission is over 200 Gbps. Currently, the Tier-1 has collected log data from 1,197 machines. 

The INFN-CNAF provides a computing farming that accounts for all computing services of the Tier-1 Bologna. It acts as a service underlying the workload management system, allowing job scheduling to access directly the INFN-CNAF experiment data. On average, about 100 thousand batch jobs are executed per day at INFN-CNAF. The resources are continuously available, 24 hours a day, 7 days a week. The CC facility is based on a warehouse infrastructure for both storage and data transfer through a distributed system \cite{Minarini}.

As Tier-1 Bologna CC concerns a dedicated infrastructure to support physics experiments \cite{chep}, minimising the resources to maintain system operationality is needed as log-data handling is a highly time and resource-consuming task. To achieve such computational cost minimisation, an approach is to identify which pieces of log data have processing priority aiming at maximising the likelihood to find useful information to the system maintenance. The present study addresses anomaly detection of the system behavior as an optimisation approach for predictive maintenance. 

The rest of this paper is structured as follows. Section~\ref{sec:sa} presents related literature on anomaly detection and system maintenance at computing centers. Section~\ref{sec:sluc} describes an evolving fuzzy-rule-based classification framework that is able to learn from summaries of log records and keep an updated representation of the spatial-temporal patterns related to the generation of log files. Section~\ref{sec:met} shows the methodology to perform the computational experiments. Classification results are given in Section~\ref{sec:er}. Conclusions are outlined in Section~\ref{sec:cfw}.

\section{Related Literature}
\label{sec:sa}
Because of the High-Luminosity LHC (HL-LHC) project, the major programmed upgrade at CERN, the used luminosity will increase by a factor of 10 from the original design. The luminosity is the rate of potential collisions per surface unit, which is proportional to the generated experimental data \cite{luminosity}. In this way, the amount of experimental and Monte Carlo analysis data will enlarge by, at least, the same factor, intensifying the maintenance complexity to keep the computing center quality of service (QoS).  

For that reason, many efforts are being done at Tier-1 Bologna in order to create predictive maintenance tools using the log data. A first work based on the Elastic Stack Suite catalogues the log records and anomalies using an embedded unsupervised ML tool \cite{Diotalevi}. Another initiative uses supervised ML approaches to predict anomalies of system behavior in an ad-hoc solution \cite{Giommi}. Another work, also focused on a content-processing strategy, provides a clustering method used to characterize log records using Levenshtein distance \cite{Rossi}. In particular, it was created a prototype to identify a normal and an anomalous system behavior, in a binary classification, considering the log data generation rate and an One-class Support Vector Machine approach \cite{Minarini}.

\subsection{The StoRM logs use case}
%\label{sec:sluc}

StoRM is the storage resource manager service (SRM) for generic disk-based storage systems adopted by the Tier-1 Bologna, providing high performance to parallel file systems.

SRM has a modular architecture made by two stateless components, Front-end (FE) and Back-end (BE), connected to database systems. The FE module manages user authentication, stores/retrieves database requests, interacting with the BE module \cite{storm}. 

\begin{figure}[htp!]
    \begin{center}
       \includegraphics[width=3.6in]{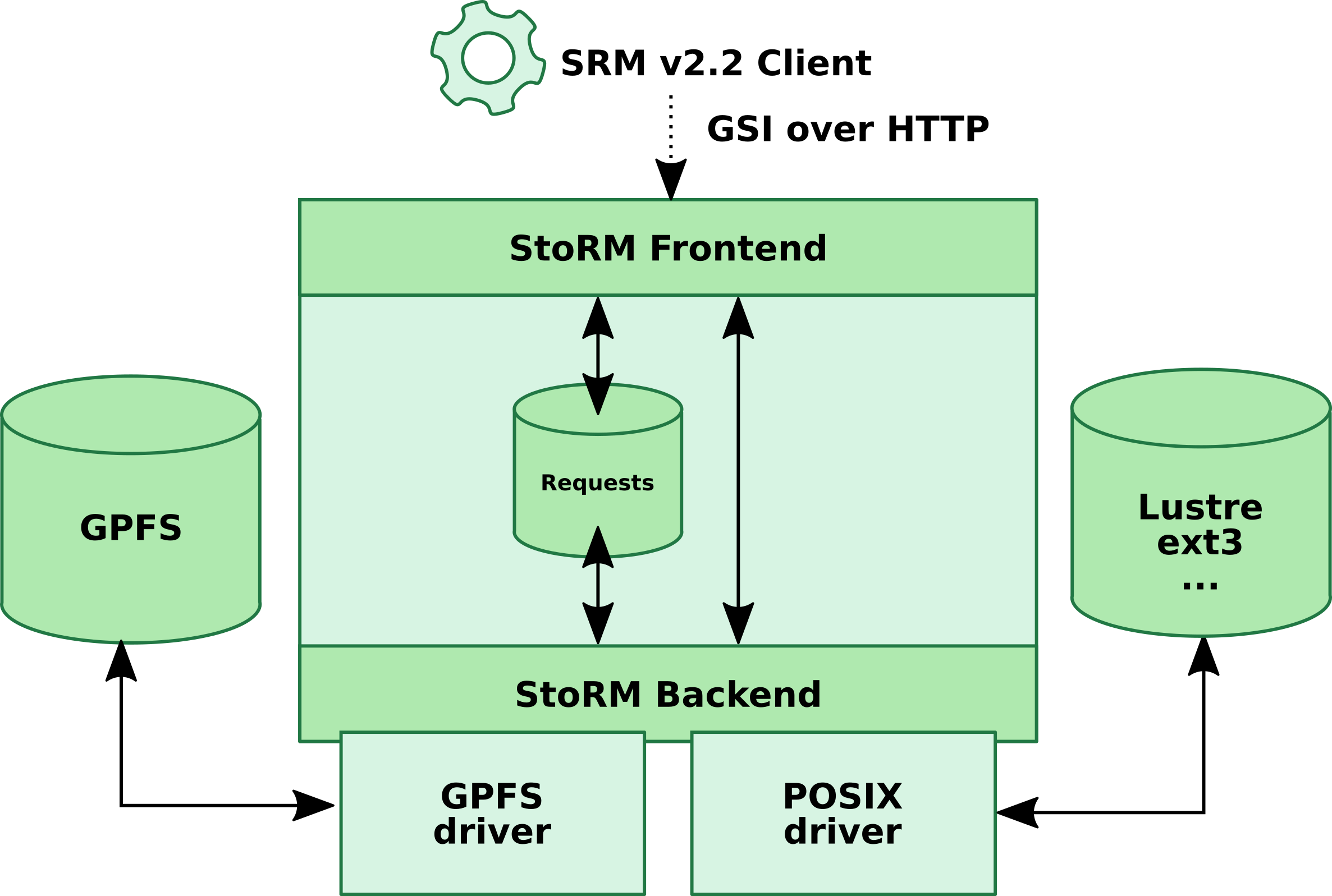}
    \end{center}
	\caption{A typical StoRM service architecture, with single back and front-end modules}
	\label{fig1}
\end{figure}

\begin{figure*}[htp!]
    \begin{center}
       \includegraphics[width=\textwidth]{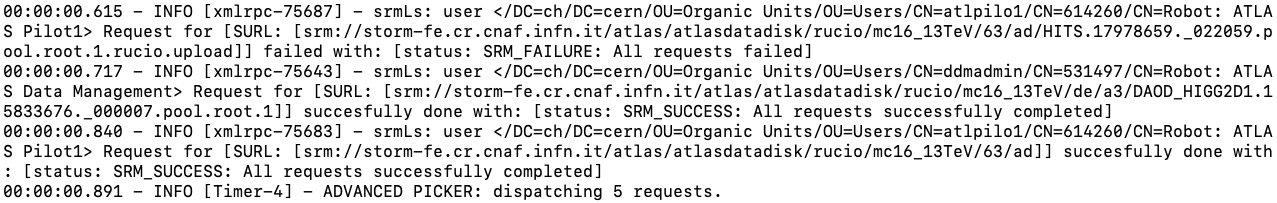}
    \end{center}
	\caption{Example of content of a \textit{storm-backend.log} file}
	\label{fig3}
\end{figure*}

In the other hand, the BE module is the core of StoRM service, executing all synchronous and asynchronous SRM functionalities and managing the Grid interactions. A simple StoRM architecture schema is presented in Fig.~\ref{fig1}, showing the main module interactions. Typically, BE log files entries include the operator that has requested the action (DN), the involved files locations (SURL) and the result of the operation. A sample of its log messages is shown in Fig.~\ref{fig3}. 

In addition, the StoRM service at Tier-1 Bologna is used by high-energy physics experiments, in which each one has a different implementation of the structures and rules of the logging. In this work, the BE module log files from ATLAS implementation is chosen as input without any special reason.

\section{eGFC: Evolving Gaussian Fuzzy Classifier}
\label{sec:sluc}

This section outlines eGFC, a semi-supervised evolving classifier derived from an online granular-computing framework \cite{Leite10} \cite{FUZZlog2}. Although eGFC handles partially labeled data, we assume a fully-labeled log-file dataset in this paper. eGFC employs Gaussian membership functions to cover the data space with fuzzy granules, and associate new data samples to class labels. Granules are scattered in the data space wherever needed to represent local information. eGFC global response comes from the aggregation of local models. A recursive algorithm constructs a rule base, and updates local models to deal with changes. eGFC addresses issues such as unlimited amounts of data and scalability \cite{Skrjanc1} \cite{Leite11}.

\subsection{Preliminaries}
\label{sec:preliminaries}

Local models are created if the newest data are sufficiently different from the current knowledge. The learning algorithm expand, reduce, delete, and merge information granules. Rules are reviewed according to inter-granular relations. eGFC provides nonlinear, non-stationary, and fuzzy discrimination boundaries among classes \cite{Skrjanc1} \cite{Leite10}. This paper particularly addresses a 4-class log-file classification problem.

Formally, let an input-output pair $(\textbf{x}, y)$ be related through $y = f(\textbf{x})$. We seek an approximation to $f$ to estimate the value of $y$ given $\textbf{x}$. In classification, $y$ is a class label, a value in a set $\{C_1, ..., C_m\} \in \mathbb{N}^m $, and $f$ specifies class boundaries. In the more general, semi-supervised case, $C_k$ may or may not be known when $\textbf{x}$ arrives. Classification of never-ending data streams involves pairs $(\textbf{x}, C)^{[h]}$ of time-sequenced data, indexed by $h$. Non-stationarity requires evolving classifiers to identify time-varying relations $f^{[h]}$.

\subsection{Gaussian Functions and Rule Structure}
\label{sec:GaussFunc}
Learning in eGFC does not require initial rules. Rules are created and dynamically updated depending on the behavior of a system over time. When a data sample is available, a decision procedure may add a rule to the model structure or update the parameters of a rule.

In eGFC models, a rule $R^i$ is

\vspace{7pt}

~~~~ IF $(x_1 $ is $ A_1^i)$ AND ... AND $(x_n $ is $ A_n^i)$

~~~~ THEN $(y $ is $ C^i)$ 

\vspace{7pt}

\noindent in which $x_j$, $j = 1, ..., n$, are attributes, and $y$ is a class. The data stream is denoted ${(\textbf{x}, y)}^{[h]}, h = 1, ...$ Moreover, $A_j^i$,  $\forall j$; $i = 1, ..., c$, are Gaussian membership functions built from the available data; and $C^i$ is the class label of the $i$-th rule. Rules $R^i$, $\forall i$, form the rule base. The number of rules, $c$, is variable, which is a notable characteristic of the approach since guesses on how many data partitions exist are needless \cite{Skrjanc1}.

A normal Gaussian membership function, $A_j^i = G(\mu_j^i, \sigma_j^i)$, has height 1 \cite{Pedrycz}. It is characterized by the modal value $\mu_j^i$ and dispersion $\sigma_j^i$. Characteristics that make Gaussians appropriate include: (i) easiness of learning and changing, i.e., modal values and dispersions can be captured and updated straightforwardly from a data stream; (ii) infinite support, i.e., since the data are priorly unknown, the support of Gaussians extends to the whole domain; and (iii) smooth surface of fuzzy granules, $\gamma^i = A^i_1 \times ... \times A^i_j \times ... \times A^i_n$, in the $n$-dimensional Cartesian space -- obtained by the cylindrical extension of uni-dimensional Gaussians, and the use of the minimum T-norm aggregation \cite{Leite11} \cite{Pedrycz}. 

\subsection{Adding Rules to the Evolving Fuzzy Classifier}
\label{sec:AddRule}
Rules may not exist \textit{a priori}. They are created and evolved as data are available. A new granule $\gamma^{c+1}$ and the rule $R^{c+1}$ that governs the granule are created if none of the existing rules $\{R^1, ..., R^c\}$ are sufficiently activated for a sample $\textbf{x}^{[h]}$. The learning algorithm assumes that $\textbf{x}^{[h]}$ brings new information. Let $\rho^{[h]} \in [0,1]$ be an adaptive threshold that determines if a new rule is needed. If
\vspace{-2pt}
\begin{eqnarray}
T\left(A_1^i(x_1^{[h]}),...,A_n^i(x_n^{[h]})\right)\leq \rho^{[h]}, ~ \forall i, ~ i = 1, ..., c, \label{activ}
\end{eqnarray}

\noindent in which $T$ is any triangular norm, then the eGFC structure is expanded. The minimum (G{\"o}del) T-norm is used in this paper, but other choices are possible. If $\rho^{[h]}$ is equal to 0, then the model is structurally stable, and unable to capture concept shifts. In contrast, if $\rho^{[h]}$ is equal to 1, eGFC creates a rule for each new sample, which is not practical. Structural and parametric adaptability are balanced for intermediate values of $\rho^{[h]}$ (stability-plasticity trade-off) \cite{Leite12}.

The value of $\rho^{[h]}$ is crucial to regulate how large granules can be. Different choices impact the accuracy and compactness of a model, resulting in different granular perspectives of the same problem. Section \ref{sec:dispersion} gives a Gaussian-dispersion-based procedure to update $\rho^{[h]}$.

A new granule $\gamma^{c+1}$ is initially represented by membership functions, $A_j^{c + 1}$, $j = 1, ..., n$, with
\vspace{-2pt}
\begin{eqnarray}
\mu_j^{c+1} = x_j^{[h]}, \label{eq6}
\end{eqnarray}

\noindent and 
\vspace{-2pt}
\begin{eqnarray}
\sigma_j^{c+1} = 1/2\pi. \label{eq7}
\end{eqnarray}

\noindent We call \eqref{eq7} the Stigler approach to standard Gaussian functions, or \textit{maximum approach} \cite{Leite11}. The intuition is to start big, and let the dispersions gradually shrink when new samples activate the same granule. This strategy is appealing for a compact model structure.

In general, the class $C^{c+1}$ of the rule $R^{c+1}$  is initially undefined, i.e., the $(c+1)$-th rule remains unlabeled until a label is provided. If the corresponding output, $y^{[h]}$, associated to $\textbf{x}^{[h]}$, becomes available, then
\vspace{-2pt}
\begin{eqnarray}
C^{c+1} = y^{[h]}. \label{eq8}
\end{eqnarray}

\noindent Otherwise, the first labeled sample of the data stream that arrives after the $h$-th time step, and activates the rule $R^{c+1}$ according to \eqref{activ}, is used to define its class, $C^{c+1}$.

In case a labeled sample activates a rule that is already labeled, but the sample's label is different from that of the rule, then a new (partially overlapped) granule and a rule are created to represent new information. Partially overlapped Gaussian granules tagged with different labels tend to have their dispersions reduced over time by the parameter adaptation procedure described in Section \ref{sec:ParAdapt}. The modal values of the Gaussian granules may also drift, if convenient for a more suitable decision boundary.

With this initial rule parameterization, preference is given to the design of granules balanced along its dimensions, rather than granules with unbalanced geometry. eGFC realizes the principle of the balanced information granularity \cite{Bargiela}, but allows the Gaussians to find more appropriate places and dispersions through adaptation mechanisms.

\subsection{Incremental Parameter Adaptation}
\label{sec:ParAdapt}
Updating the eGFC model consists in: (i) reducing or expanding Gaussians $A_j^{i^*}$, $\forall j$, of the most active granule, $\gamma^{i^*}$, considering labeled and unlabeled samples; (ii) moving granules toward regions of relatively dense population; and (iii) tagging rules if labeled data are available. Adaptation aims to develop more specific local models in the sense of Yager \cite{Yager}, and provide pavement (covering) to the newest data.

A rule $R^i$ is candidate to be updated if it is sufficiently activated by an unlabeled sample, $\textbf{x}^{[h]}$, according to
\vspace{-2pt}
\begin{eqnarray}
min\left(A_1^i(x_1^{[h]}),...,A_n^i(x_n^{[h]})\right) > \rho^{[h]}. \label{eq9}
\end{eqnarray} 

\noindent Geometrically, $\textbf{x}^{[h]}$ belongs to a region highly influenced by the granule $\gamma^i$. Only the most active rule, $R^{i^*}$, is chosen for adaptation in case two or more rules reach the $\rho^{[h]}$ level for the unlabeled $\textbf{x}^{[h]}$. For a labeled sample, i.e., for pairs $(\textbf{x},y)^{[h]}$, the class of the most active rule $R^{i^*}$, if defined, must match $y^{[h]}$. Otherwise, the second most active rule among those that reached the $\rho^{[h]}$ level is chosen for adaptation, and so on. If none of the rules are apt, then a new one is created (Section \ref{sec:AddRule}).

To include $\textbf{x}^{[h]}$ in $R^{i^*}$, eGFC's learning algorithm updates the modal values and dispersions of the corresponding membership functions $A_j^{i^*}$, $j = 1, ..., n$, from
\vspace{-2pt}
\begin{eqnarray}
\mu_j^{i^*}(new) = \frac {(\varpi^{i^*}-1) \mu_j^{i^*}(old) + x_j^{[h]}}{\varpi^{i^*}}, \label{eq10}
\end{eqnarray} 

\noindent and
\vspace{-2pt}
\begin{eqnarray}
\sigma_j^{i^*}(new) &=& \biggl( \frac {(\varpi^{i^*}-1)}{\varpi^{i^*}} ~ \left(\sigma_j^{i^*}(old)\right)^2 ~+ \nonumber \\
&& +~ \frac {1}{\varpi^{i^*}} \left(x_j^{[h]} - \mu_j^{i^*}(old)\right)^2 \biggr)^{1/2}, ~~~ \label{eq11}
\end{eqnarray} 

\noindent in which $\varpi^{i^*}$ is the number of times the $i^*-th$ rule was chosen to be updated. Notice that \eqref{eq10}-\eqref{eq11} are recursive and, therefore, do not require data storage. As $\sigma^{i^*}$ defines a convex region of influence around $\mu^{i^*}$, very large and very small values may induce, respectively, a unique or too many information granules per class. An approach is to keep $\sigma_j^{i^*}$ between a lower, $1/4\pi$, and the Stiegler, $1/2\pi$, limits.

\subsection{Dispersion-Based Time-Varying $\rho$-Level}
\label{sec:dispersion}
Let the activation threshold, $\rho^{[h]} \in [0,1]$, be time-varying, similar to \cite{Leite10} \cite{Garcia}. The threshold assumes values in the unit interval according to the overall average dispersion
\vspace{-2pt}
\begin{eqnarray}
\sigma_{avg}^{[h]} &=& \frac{1}{cn} \sum\limits_{i=1}^c \sum\limits_{j=1}^n \sigma^{i[h]}_j, \label{eq12}
\end{eqnarray} 

\noindent in which $c$ and $n$ are the number of rules and attributes, so that
\vspace{-2pt}
\begin{eqnarray}
\rho(new) &=& \frac{\sigma_{avg}^{[h]}}{\sigma_{avg}^{[h-1]}} ~ \rho(old). \label{eq13}
\end{eqnarray} 

As mentioned, rules' activation levels for an input $\textbf{x}^{[h]}$ are compared to $\rho^{[h]}$ to decide between parametric or structural changes of an eGFC model. In general, eGFC starts learning from an empty rule base, and without knowledge about the properties of the data. Practice suggests $\rho^{[0]} = 0.1$ as starting value. The threshold tends to converge to a proper value after some time steps if the classifier structure and parameters achieve a level of maturity and stability. Non-stationarities and new classes guide $\rho^{[h]}$ to values that better reflect the needs of the current environment. A time-varying $\rho^{[h]}$ avoids assumptions about how often the data stream changes.

\subsection{Merging Similar Granules}
\label{sec:mergGranules}
Similarity between two granules with the same class label may be high enough to form a unique granule that inherits the essential information conveyed by the merged granules. Analysis of inter-granular relations requires a distance measure between Gaussian objects. Let
\vspace{-2pt}
\begin{eqnarray}
d(\gamma^{i_1}, \gamma^{i_2}) &=& \frac {1}{n} \biggl( ~\sum_{j=1}^n | \mu_j^{i_1} - \mu_j^{i_2} | + \sigma_j^{i_1} +  \nonumber \\ && \sigma_j^{i_2} - 2 \sqrt{\sigma_j^{i_1} \sigma_j^{i_2}} ~\biggr)  \label{eq14}
\end{eqnarray}

\noindent be the distance between the granules $\gamma^{i_1}$ and $\gamma^{i_2}$. This measure considers Gaussians and the specificity of information, that is, in turn, inversely related to the Gaussians' dispersion \cite{Soares}. For example, if the dispersions $\sigma_j^{i_1}$ and $\sigma_j^{i_2}$ differ one from another, rather than being equal, the distance between the underlying Gaussians is larger.

eGFC may merge the pair of granules that presents the smallest value of $d(.)$ for all pairs of granules. Both granules must be either unlabeled or tagged with the same class label. The merging decision is based on a threshold value, $\Delta$, or expert judgement regarding the suitability of combining such granules to have a more compact model. For data within the unit hyper-cube, we suggest $\Delta = 0.1$ as default, which means that the candidate granules should be quite similar and, in fact, carry the same information.

A new granule, say $\gamma^i$, which results from $\gamma^{i_1}$ and $\gamma^{i_2}$, is built by Gaussians with modal values
\vspace{-2pt}
\begin{eqnarray}
\mu_j^i = \frac{\frac{\sigma_j^{i_1}}{\sigma_j^{i_2}}\mu_j^{i_1} + \frac{\sigma_j^{i_2}}{\sigma_j^{i_1}}\mu_j^{i_2}}{\frac{\sigma_j^{i_1}}{\sigma_j^{i_2}} + \frac{\sigma_j^{i_2}}{\sigma_j^{i_1}}}, ~  j=1,...,n,  \label{eq15}
\end{eqnarray}

\noindent and dispersion
\vspace{-2pt}
\begin{eqnarray}
\sigma_j^i = \sigma_j^{i_1} + \sigma_j^{i_2}, ~ j=1,...,n. \label{eq16}
\end{eqnarray}

\noindent These relations take into consideration the uncertainty ratio of the original granules to determine an appropriate location and size of the resulting granule. Merging granules reduces the number of rules and redundancy \cite{Leite10} \cite{Soares}.

\subsection{Deleting Rules}
\label{sec:delRules}
A rule is removed from the eGFC model if it is inconsistent with the current environment. In other words, if a rule is not activated for a number of iterations, say $h_r$, then it is deleted from the rule base. However, if a class is rare, then it may be the case to set $h_r$ to infinity and keep the inactive rules. Removing rules periodically helps to keep the knowledge base updated in some applications.

\subsection{Semi-Supervised Learning from Data Streams}
\label{sec:learning}
The semi-supervised learning procedure to construct and update eGFC models along their lifespan is given in the next column.

~~

\section{Methodology}
\label{sec:met}

We describe a dynamic control chart approach we propose for attribute extraction and log data tagging. We give details about the data-set and evaluation measures.

\subsection{Control Chart: Tagging Log Data}

A control chart is a time-series graph to monitor the evolution of a process, phenomenon or variable based on the Central Limit theorem \cite{controlChart}. The main idea is that the mean, $\mu(u)$, of an independent random variable, $u$, with unknown distribution, follows a normal distribution. 

Let 
\vspace{-2pt}
\begin{eqnarray}
\textbf{u}_j = [u_1 ~ \dots ~ u_i ~ \dots ~ u_n],
\label{eq17}
\end{eqnarray}

\noindent be a sequence of values that represent the number of log entries in a log file over a time window $w_j = [\underline{w}_j ~  \overline{w}_j]$; $\textbf{u}_j \in \mathbb{N}^n$. The time interval from $u_1$ to $u_n$ coincides with the window boundaries, $\underline{w}_j$ and  $\overline{w}_j$. Additionally, let $\mu_j$ be the mean of $\textbf{u}_j$, thus
\vspace{-2pt}
\begin{eqnarray}
\mu_j = \frac{1}{n} \sum\limits_{i=1}^n u_i, ~ u_i \in [\underline{w}_j ~  \overline{w}_j].
\label{eq18}
\end{eqnarray}

A time series of means, with cardinality $m$, is
\vspace{-2pt}
\begin{eqnarray}
\mu = [\mu_1 ~ \dots ~ \mu_j ~ \dots ~ \mu_m].
\label{eq19}
\end{eqnarray}

\begin{figure}[htp!]
    \begin{center}
       \includegraphics[width=3.4in]{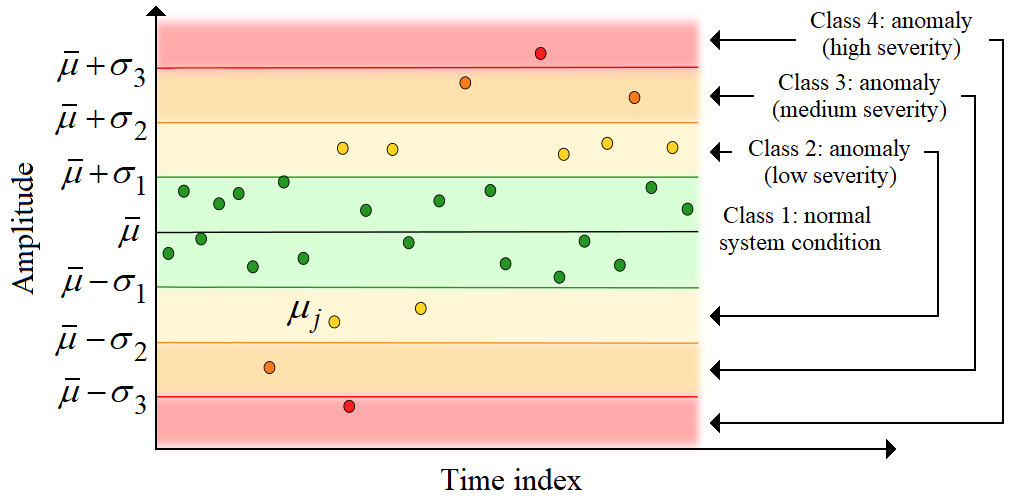}
    \end{center}
	\caption{Control chart used to tag mean log data within a time window}
	\label{fig4}
\end{figure}

\noindent As $\mu$ follows a normal distribution, a sample $\mu_j$ can be tagged by means of a control chart, see Fig.~\ref{fig4}. The mean of the time

\newpage

\hrule
\vspace{6pt}
\textbf{eGFC: Online Semi-Supervised Learning}
\vspace{3pt}
\hrule
\vspace{4pt}
\begin{algorithmic}[1]
	\STATE Initial number of rules, $c = 0$;
	\STATE Initial meta-parameters, $\rho^{[0]} = \Delta = 0.1$, $h_r = 200$;
    \STATE Read input data sample $\textbf{x}^{[h]}, h=1$;
    \STATE Create granule $\gamma^{c+1}$ (Eqs. \eqref{eq6}-\eqref{eq7}), unknown class $C^{c+1}$;
    \STATE \textbf{FOR} $h$ = 2, ... \textbf{DO}
    \STATE ~~ Read $\textbf{x}^{[h]}$;
    \STATE ~~ Calculate rules' activation degree (Eq. \eqref{activ});
    \STATE ~~ Determine the most active rule $R^{i^*}$;
    \STATE ~~ Provide estimated class $C^{i^*}$;
    \STATE ~~ // Model adaptation
    \STATE ~~ \textbf{IF} $T(A^i_1(x_1^{[h]}), ..., A^i_n(x_n^{[h]})) \leq \rho^{[h]} ~ \forall i, ~ i = 1, ..., c$
    \STATE ~~~~ \textbf{IF} actual label $y^{[h]}$ is available
    \STATE ~~~~~~ Create labeled granule $\gamma^{c+1}$ (Eqs. \eqref{eq6}-\eqref{eq8});
    \STATE ~~~~ \textbf{ELSE}
    \STATE ~~~~~~ Create unlabeled granule $\gamma^{c+1}$ (Eqs. \eqref{eq6}-\eqref{eq7});
    \STATE ~~~~ \textbf{END}
    \STATE ~~ \textbf{ELSE}
    \STATE ~~~~ \textbf{IF} actual label $y^{[h]}$ is available
    \STATE ~~~~~~ Update the most active granule $\gamma^{i^*}$ whose class \\ 
    ~~~~~~ $C^{i^*}$ \hspace{-6pt} is equal to $y^{[h]}$ (Eqs. \eqref{eq10}-\eqref{eq11});
    \STATE ~~~~~~ Tag unlabeled active granules;
    \STATE ~~~~ \textbf{ELSE}
    \STATE ~~~~~~ Update the most active  $\gamma^{i^*}$ (Eqs. \eqref{eq10}-\eqref{eq11});
    \STATE ~~~~ \textbf{END}
    \STATE ~~ \textbf{END}
    \STATE ~~ Update the $\rho$-level (Eqs. \eqref{eq12}-\eqref{eq13});
    \STATE ~~ Delete inactive rules based on $h_r$;
    \STATE ~~ Merge granules based on $\Delta$ (Eqs. \eqref{eq14}-\eqref{eq16});
    \STATE \textbf{END}    
\end{algorithmic}
\vspace{4pt}
\hrule
\vspace{6pt}

~~

%\clearpage

\vspace{7pt}

\noindent series of means $\mu$ is
\vspace{-2pt}
\begin{eqnarray}
\overline{\mu} = \frac{1}{m} \sum\limits_{j=1}^{m} \mu_j.
\label{eq20}
\end{eqnarray}

\noindent The $k$-th upper and lower horizontal lines in relation to $\overline{\mu}$ refer to the $k$-th standard deviation,
\vspace{-2pt}
\begin{eqnarray}
\sigma_k(\mu) = k\sqrt{\frac{1}{m}\sum\limits_{j=1}^m (\overline{\mu} - \mu_j)^2},
\label{eq21}
\end{eqnarray}

\noindent such that if a sample 
\vspace{-2pt}
\begin{eqnarray}
\mu_j ~ \subset ~ [\overline{\mu} - \sigma_k(\mu_ j \forall j), ~ \overline{\mu} +  \sigma_k(\mu_ j \forall j)], \label{intclasses}
\end{eqnarray}

\noindent for $k = 1$, it is tagged as `Class 1' (normal system condition). Otherwise, if \eqref{intclasses} holds for $k = 2, 3$, and $4$, respectively, $\mu_j$ is tagged as `Class 2', `Class 3', and `Class 4', which mean low, medium, and high-severity anomaly. The greater the value of $k$, the greater the severity of the anomalous behavior.

Control charts are widely used in quality monitoring to identify anomalies according to the control lines calculated from a stream of means. The probability that a sample $\mu_j$ is within the different class boundaries are 68.3\%, 27.1\%, 4.3\%, and 0.3\%, respectively. Therefore, the online data classification problem is unbalanced.

\subsection{About the Data-set}

A stream of time-indexed log records is generated by the StoRM service. Each log entry is composed by the timestamp in which it was written, and the message itself. Analysis of the message type and its content is out of the scope of this paper. 

We extract relevant attributes from the original log data stream by analysing constant sliding time windows. Transformed data are provided as 5-attribute vectors 

\begin{eqnarray}
\textbf{x} = [x_1 ~~ x_2 ~~ x_3 ~~ x_4 ~~ x_5],
\end{eqnarray} 

\noindent whose elements evaluated in time window $ w_j$ are $\overline{\mu}$, $\sigma(\mu_ j~\forall j)$, $min(\mu_ j~\forall j)$, $max(\mu_ j~\forall j)$, and $max(\Delta \mu_j~\forall j)$. The latter means the maximum difference of the amplitude of two consecutive $\mu_j$, belonging to the time window $w_j$.

A vector $\textbf{x}^{[h]}$ is associated to a class label $C = \{1, 2, 3, 4\}$ that, in turn, indicates the system behavior. The true label $C$ is available after an estimation $\hat{C}$ provided by the eGFC model. The pair $(\textbf{x},C)^{[h]}$ is used by the eGFC online learning algorithm for an updating step.

\subsection{Performance Measure}

Classification accuracy, $Acc \in [0,1]$, is computed recursively from

\begin{eqnarray}
Acc(new) = \frac{h-1}{h} ~ Acc(old) + \frac{1}{h} ~ \tau,
\end{eqnarray}

\noindent in which $\tau := 1$ if $\hat{C}^{[h]} = C^{[h]}$ (right estimation). Otherwise, $\tau := 0$ (wrong class estimation).

The average number of granules or rules over time, $c_{avg}$, is a measure of model concision. Recursively,

\begin{eqnarray}
c_{avg}(new) = \frac{h-1}{h} ~ c_{avg}(old) + \frac{1}{h} ~ c^{[h]}.
\end{eqnarray}

%\vspace{5pt}

\section{Results}
\label{sec:er}

We evaluate the evolving Gaussian fuzzy classification system. No prior knowledge about the data is assumed. Classification models are developed from scratch based on information extracted from an online log data stream.

\subsection{eGFC Results}

We look for an evolving classifier based on the newest input data. The default meta-parameters are used (see \textit{eGFC Learning Algorithm}). Table \ref{Tab1} summarizes the results averaged over 5 runs for shuffled data-sets extracted from log records. Four data-sets were produced using the same data, but different lengths of time windows, namely 60, 30, 15, and 5-minute time windows. Larger time windows impose a higher-order low-pass filter effect, and tends to isolate the trend component of the time series from cyclical and random (stochastic) components. Each data-set consists of 1,436 samples and 5 attributes. Four classes are possible, namely `normal operation', `low severity', `medium severity', and `high severity'.

\begin{table}[!ht]
	\small \caption{eGFC Performance in Multi-class Classification of System Anomalies ($99\%$ of confidence)}
	\vspace{-15pt}
	\begin{center}
		\resizebox{\columnwidth}{!}{
			\begin{tabular}{c|ccc}
				\hline
				Lenght (min) & $Acc$(\%) & \# Rules & Time (s) \\
				\hline
				%0.59  2.10 0.05
				60 & $92.48 \pm 1.21$ & $13.42 \pm 4.32$ & $0.36 \pm 0.10$ \\
				% 2.41 1.26 0.02
				30 & $88.01 \pm 4.96$ & $17.22 \pm 2.59$ & $0.45 \pm 0.04$ \\
				% 2.74 2.33 0.05
				15 & $82.57 \pm 5.64$ & $18.13 \pm 4.79$ & $0.49 \pm 0.10$ \\ 
				% 2.44 1.22 0.03
				5 & $81.97 \pm 5.02$ & $16.09 \pm 2.51$ & $0.41 \pm 0.06$ \\
				\hline
			\end{tabular}
			\label{Tab1}}
	\end{center}
\end{table}

Table \ref{Tab1} shows that analysis of larger 60-minute windows facilitates the eGFC learning algorithm to detect and classify spatial-temporal patterns, which represent the anomaly classes. Notice that using a more compact model structure ($13.42$ fuzzy rules on average along the learning steps), the eGFC model produced an average accuracy of $92.48\%$. The CPU time in a quad-core i7-8550U with 1.80GHz and 8GB of RAM are similar in all scenarios.

\begin{figure}[htp!]
    \begin{center}
       \includegraphics[width=3.2in]{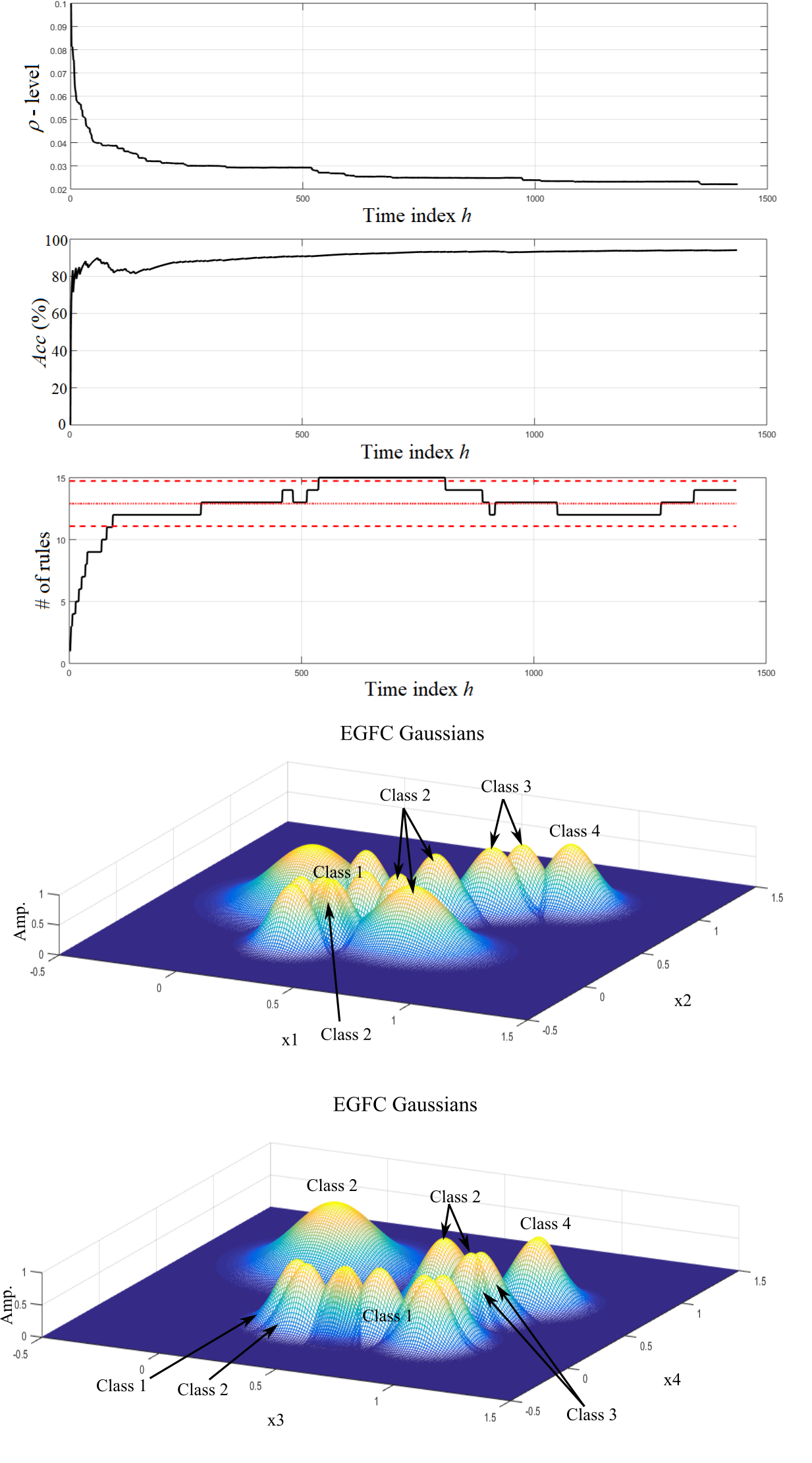}
    \end{center}
	\caption{The time evolution of the evolving factors: granulation $\rho$ and number of rules, and the model accuracy until the convergence at the 3 first graphics. At the last 2 graphics the eGFC Gaussian classes.}
	\label{hills}
\end{figure}

Figure \ref{hills} gives a typical example of evolution of the $\rho$-level, accuracy, and number of eGFC rules. Four dimensions of the final Gaussian granules, at $h = 1436$, are also shown. Notice that data from Class 2 and Class 3 (low and medium-severity anomalies) spread in a nonlinear way over the data space. These classes require more than one granule and rule to be represented, whereas the remaining classes are generally given in a common region. Class-4 data (high-severity anomaly) belong to a more compact region than the data of other classes and, therefore, are represented by a single granule. A higher number of granules to represent a class, in general, provides larger non-linearity of decision boundaries, which improves classification accuracy. 

\begin{figure}[htp!]
    \begin{center}
       \includegraphics[width=3.2in]{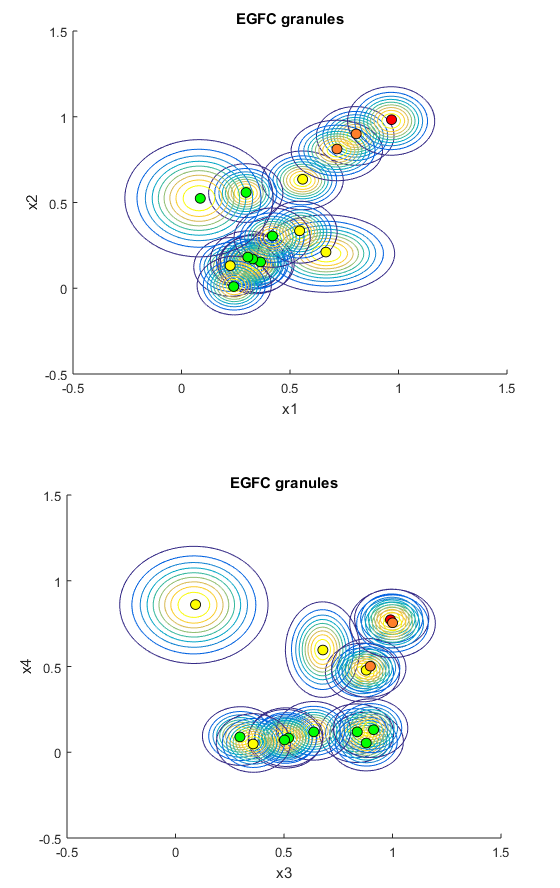}
    \end{center}
	\caption{The multi-dimensional ellipsoidal geometry of eGFC granules using the first four attributes of the log stream. The colours of the centers refer to the control chat of Fig. \ref{fig4}, i.e., green: normal system condition; yellow, orange and red: low, medium and high anomaly severity}
	\label{elipses}
\end{figure}

\begin{figure}[htp!]
    \begin{center}
       \includegraphics[width=3in]{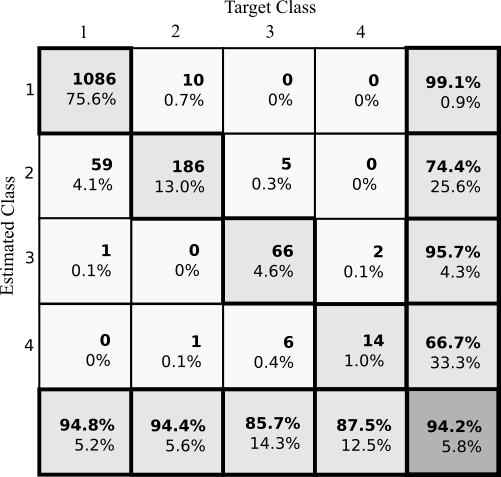}
    \end{center}
	\caption{Example of confusion matrix provided by a $94.2\%$-accuracy eGFC model}
	\label{confusionMatrix}
\end{figure}

Figure \ref{elipses} emphasizes the multi-dimensional ellipsoidal geometry of eGFC granules. This contour lines representation confirm the spreading characteristic specially related to Class-2 data, showing large overlapping regions of Class-1 and Class-2. Figure \ref{confusionMatrix} shows the confusion matrix for a $94.08\%$-accuracy scenario. Notice that confusion happens in the neighbourhood of a target class, which means that if a higher number of streaming samples are further available, the eGFC model may improve its accuracy by fine tuning its decision boundaries. Class 1 (normal operation) and Class 2 (low severity) are those responsible for a larger reduction of the overall accuracy.

To sum up, using the evolving fuzzy classification methodology and the sliding window control-chart-based approach, a CC maintenance system can accurately identify time windows that require further analysis in terms of text content. The evolving methodology supports data and information mining to assist predictive maintenance. Overall system status can be modelled as Gaussian granules of the log activity rate, and status changing can be noticed visually from the control charts. In addition, the stream of system status can be used to diagnose the context of the current log status, and to predict the next status. Since eGFC preserves its accuracy in non-stationary environment, the approach has shown to be a reliable solution to the predictive maintenance problem.

%\subsection{Performance Comparison}

%We compare the performance of the proposed eGFC model with that of ..... on the 60-minute window scenario.

%***********Final Table

%Graphics \cite{schmidt2018predictive}

\section{Conclusion}
\label{sec:cfw}
We described a real-time evolving general-purpose solution, namely, eGFC, to the log-based anomaly detection problem considering time-varying data from the Tier-1 Bologna computer center. eGFC models achieved an average accuracy of $92.48\% \pm 1.21$ with a confidence interval of $99\%$ using a 60-minute sliding time window. Since the anomaly detection issue is context-sensible, the eGFC approach provides a strategy to update and evolve information granules and the parameters and structure of a fuzzy rule-based classifier in real-time. Multi-dimensional Gaussian granules are placed and sized autonomously in the data space aiming at constant improving the classification performance.

Fuzzy information granulation gives flexible and smooth boundaries to the classification model such that a wide variety of computer-center behaviors related to the same class label -- even occurring in a conflicting region with overlapped classes -- could be captured. This way, the eGFC approach, as a data-stream-oriented method, has shown to be highly applicable to a large range of classification issues concerning large log records from computing centers such as the Tier-1 Bologna, which supports the high-energy physics experiments at the Large Hadron Collider. Additionally, the autonomous sliding-window-based tagging strategy using control charts was successfully applied to the anomaly detection problem in question. Hand-labelling large volumes of online data (a key research issue in the machine learning community) is usually infeasible. Therefore, the chart-based approach seems quite promising to lead accuracy improvement in evolving classification frameworks. 

The present study provides basis for extracting information from log content and identifying the best components to be text-processed, which minimise computational resource consumption. In the future we shall identify the type of message associated to anomalous time windows, and investigate autonomous feature extraction procedures.

\bibliography{biblio}
%ieeetr plainnat
\bibliographystyle{ieeetr}

\end{document}